# A CMOS Spiking Neuron for Dense Memristor-Synapse Connectivity for Brain-Inspired Computing

Xinyu Wu, Vishal Saxena, Kehan Zhu
Department of Electrical and Computer Engineering
Boise State University
Boise, Idaho, United States
xinyuwu@u.boisestate.edu

*Abstract* — Neuromorphic systems that densely integrate CMOS spiking neurons and nano-scale memristor synapses open a new avenue of brain-inspired computing. Existing silicon neurons have molded neural biophysical dynamics but are incompatible with memristor synapses, or used extra training circuitry thus eliminating much of the density advantages gained by using memristors, or were energy-inefficient. Here we describe a novel CMOS spiking leaky integrate-and-fire neuron circuit. Building on a reconfigurable architecture with a single opamp, the described neuron accommodates a large number of memristor synapses, and enables online spike timing dependent plasticity (STDP) learning with optimized power consumption. Simulation results of an 180nm CMOS design showed 97% power efficiency metric when realizing STDP learning in 10,000 memristor synapses with a nominal 1MΩ memristance, and only 13μA current consumption when integrating input spikes. Therefore, the described CMOS neuron contributes a generalized building block for large-scale brain-inspired neuromorphic systems.

## I. INTRODUCTION

Brain-inspired computing is an emerging paradigm, spurred by advances in more understanding of biological spiking neural networks (SNNs) and nano-scale memristive devices invented as minuscule electrical synapses. By exploiting memristor synapses integrated on a standard CMOS chip, it is conceivable to build neuromorphic very large-scale integration (VLSI) systems that mimic the computation occurring in a brain cortex [1]–[4]. Neuromorphic computing architectures are promising candidates to address the challenges of energy-efficiency and restricted parallelism associated with the conventional von Neumann computing architectures. To this end, energy-efficient spiking silicon neuron circuits are needed as fundamental building blocks for realizing these systems.

Since the emergence of nano-scale memristors, there has been a growing interest in integrating these memristor synapses with CMOS neurons to realize novel neuromorphic functionality. These conceptual implementations intend to exploit the spike-timing-dependent-plasticity (STDP) learning property of memristor devices to realize machine learning in hardware [2]–[20]. In these approaches, researchers have used compact leaky integrate-and-fire neuron (IFN) circuits as abstraction for the biological neuron that has reasonable accuracy to be useful for neural learning and need a far lower number of transistors to implement. Fig. 1 illustrates a crossbar organization of such SNNs using IFNs with memristor synapses. The synapse weights are locally updated using the STDP rule where the change in weight depends upon the relative firing times of the pre- and post-synaptic neurons. However, existing IFN designs have focused on modelling a certain aspect of neural dynamics but rejected memristor synapses [21]–[24], or need extra learning circuitry thus eliminating much of the density advantages gained by using memristors [11], or were energy-inefficient for larger memristive network [25]–[27].

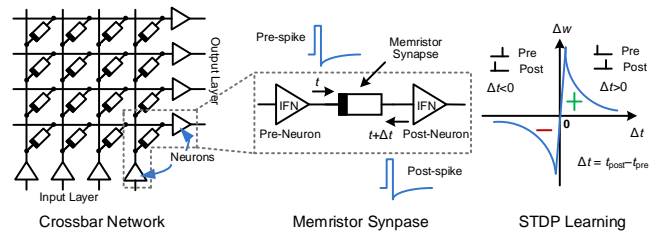

Fig. 1. Crossbar SNN architecture with memristor synapses, a synapse connected between two spiking neurons showing pre- and post-synaptic spikes, and graphical depiction of a bio-inspired pair-wise STDP-learning rule.

In this paper, a novel CMOS spiking IFN circuit is proposed. It assembles a biological plausible spike generator in a reconfigurable architecture with dynamically biased single opamp. With an innovative dual-mode operation, the proposed neuron works like a two-terminal block with respect to memristor synapses, thus enables online STDP learning and provides large driving capability to accommodate thousands memristors in parallel during firing while consumes a very low power during integration. The proposed neuron was implemented in an 180nm CMOS process. Simulation results verified its functionality as the generalized building blocks together with the two-terminal memristor synapse to form a simple repeating structure in the same way as biological neural systems. Using a device model [28] fitted to existing memristors [28]–[34], simulations showed 97% power efficiency when driving STDP learning in 10,000 memristor synapses with average 1MΩ memristance, and 13µA current consumption during integration mode. Therefore, it is amendable to scale-up for large-scale neuromorphic systems required for brain-inspired computing.

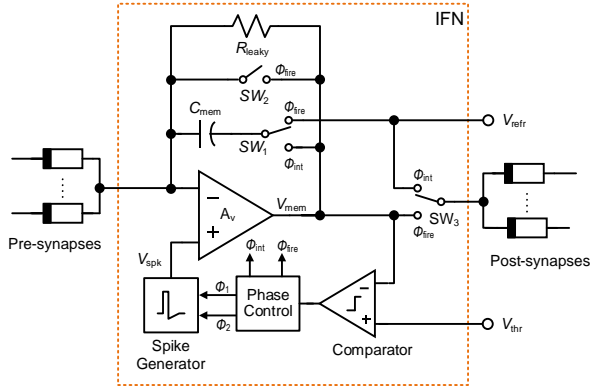

Fig. 3. Diagram of the proposed novel leaky integrate-and-fire neuron circuit. The reconfigurable architecture makes it like a simple two-terminal building block for memristor synapses, while enables current summing, large current driving, and online STDP learning with a single opamp.

## II. SPIKING NEURON CIRCUIT

As previously discussed, IFN circuits were used to emulate large-scale spiking neural networks because they offer reasonable accuracy to neural learning and compact silicon implementation. The IFNs generate spikes with the desired action potential (or spike waveform), and drive the memristor synapses with pre- and post-synaptic potentials. However, existing IFN circuits suffer several problems and are difficult to fit into large-scale neuromorphic systems with memristor synapses.

Firstly, to integrate currents across memristor synapses (e.g. 100kΩ to 100MΩ resistance range) and drive thousands of these in parallel, the conventional current-input IFN architecture [3] cannot be directly employed: current summing overheads and the large current drive required from the neurons would be prohibitive. Instead, an opamp-based IFN is desirable as it provides the required current summing node and a large current drive capability. Further, large current drive capability generally resulting large power consumption. Simply using an opamp to drive many memristors generally yielded energy-inefficient IFN designs, therefore preventing scale-up [25]–[27].

Secondly, conventional IFN circuits were designed to generate spikes to match spiking behaviors of certain biological neurons [21], and then, synapse learning is barely taken into consideration together with the neuron circuit. However, brain-inspired STDP learning in memristor synapse requires the neuron to produce spikes, or action potentials, with specific shape [4]. Therefore, to realize online learning, a pulse generator is needed to produce STDP-type spikes that are compatible with the electrical properties of the two-terminal memristors. Moreover, a configurable STDP spike shape is desired to enable the designed silicon neuron to deal with a variety of memristor devices and incorporate spike-based learning algorithms, both of which are continuously evolving.

Finally, the primary benefit to use nano-scale memristor as synapse is its high integration capability that is ideal for the implementation of a huge number of synapses. For this reason, any accessory circuitry attached to synapse for online learning neutralize this benefit and even can make memristor synapse less desirable if the accessory circuitry is big. Thus, the simplest single wire connection between a synapse to a neuron is expected. To get rid of accessory circuits, current summing and pre-spike driving should be implemented on the same node, and post-spike propagating and large current driving are required to implement on another same node as well. So, a compact neuron architecture utilizing opamp driver for both pre- and post-spikes is expected.

Fig. 2 shows the circuit schematic of the proposed leaky integrate-and-fire neuron. It is composite of a single-ended opamp, an asynchronous comparator, a phase controller, a spike generator, three analog switches ($SW_1$, $SW_2$ and $SW_3$), a capacitor $C_{mem}$ for integration operation, and a leaky resistor $R_{leaky}$ that is implemented using a MOS transistor in triode. Its dual-mode operation and STDP-compatible spike generation is the key to overcome three challenges discussed before.

### A. Dual-mode Operation

Dual-mode operation uses single opamp as both an integrator as well as the driving buffer. Here, a power-optimized opamp operates in two asynchronous modes: integration and firing modes, as illustrated in Fig. 3.

In integration mode, phase control signal $\Phi_{int}$ is set to active (logic high), and switch $SW_1$ is set to connect "membrane" capacitor $C_{mem}$ with the output of the opamp. Because phase control signal $\Phi_{fire}$ is complementary to $\Phi_{int}$, switch $SW_2$ and switch $SW_3$ that connects to post-synapses are both open. Thanks to the spike generator that is designed to hold to the refractory potential ($V_{refr}$) during the non-firing time, the positive port of opamp is set to voltage $V_{refr}$, which in fact acts as the common mode voltage. With this configuration, the opamp realizes a leaky integrator with the leak-rate controlled by the triode transistor $R_{leaky}$, and charges the capacitor $C_{mem}$ resulting in the neuron "membrane potential" $V_{mem}$. Now, the neuron sums currents flow into it and causes $V_{mem}$ to move down, noting that this is a negative integrator. Then $V_{mem}$ is compared with a threshold $V_{thr}$, crossing which triggers the spike-generation circuit and forces the opamp into the "firing phase".

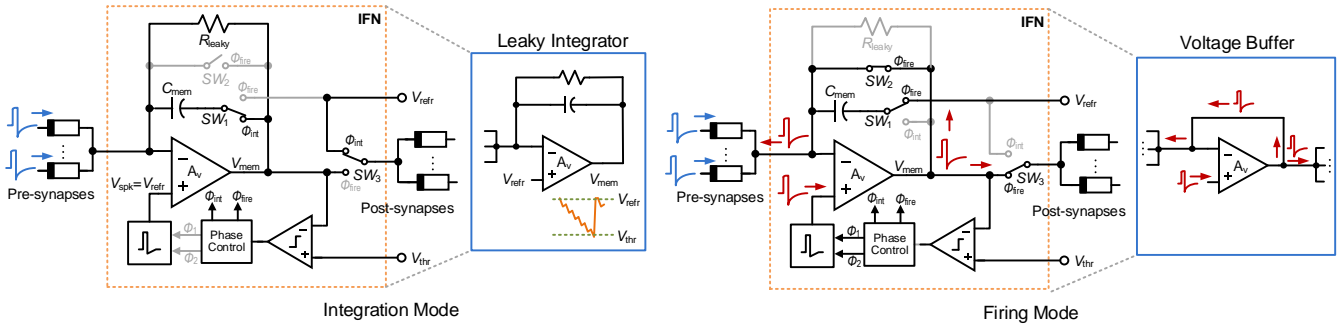

Fig. 2. Dual-mode operation of the proposed leaky integrate-and-fire neuron (a) Integration mode: Opamp is configured as a leaky integrator to sum up currents flow into neuron; (b) Firing mode: Opamp is reconfigured as a voltage buffer to drive memristor synapses with the spiking action potential.

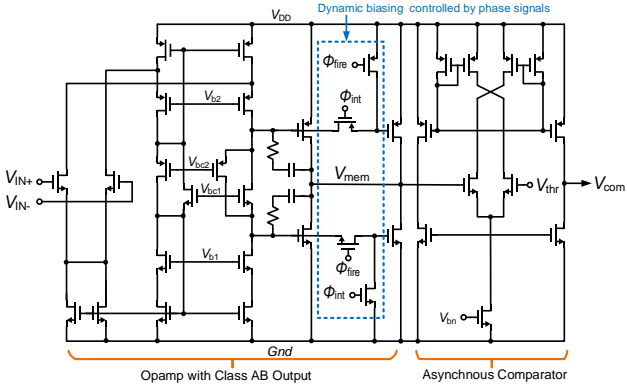

Fig. 4. Circuit details of the embedded opamp and comparator. A dynamic biased class-AB stage optimizes power consumption of opamp by using a large drive current only in firing mode.

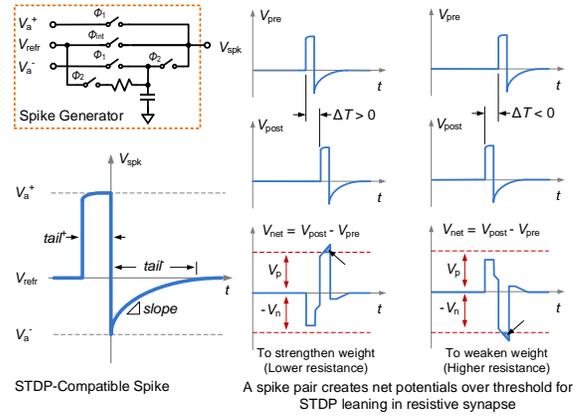

Fig. 5. STDP-compatible spike generation with tunable parameters. These spikes run across a memristor synapse and reduce resistance if $V_{net} > V_p$, or increase resistance if $V_{net} < -V_n$.

During the firing-phase, phase signals $\Phi_{fire}$ is set to active (logic high) and $\Phi_{int}$ is set to inactive (logic low) which causes switch $SW_2$ is close and switch $SW_3$ bridges opamp output to post-synapses. Now, the opamp is reconfigured as a voltage buffer. The STDP spike generator creates the required action potential waveform $V_{spk}$ (will be discussed later) and sends to input port of the buffer, which is the positive port of the opamp. Noting that both pre-synapses and post-synapses are shorted to the buffer output, the neuron propagates post-synaptic spikes in backward direction on the same port as that of current summing, and pre-synaptic spikes in forward direction on the same node of post-synapse driving. At the same time, $SW_1$ is connected to $V_{refr}$, and then discharges the capacitor $C_{mem}$.

For circuit realization, we use a folded-cascode opamp with a split dynamically biased class-AB output stage. For optimum energy consumption, the main branch of the class-AB stage is shut-off during integration mode under the control of phase signals $\Phi_{int}$ and $\Phi_{fire}$; during firing mode, it is turned-on and provides the required ability of large current driving. A dedicated asynchronous comparator is used to compare neuron membrane potential against the firing threshold. To accommodate the STDP learning, comparator hysteresis was traded-off with the speed. Fast transient response is desired to create significant STDP learning. A basing circuitry provides $V_{b1}$, $V_{b2}$, $V_{bc1}$, $V_{bc2}$, and $V_{bn}$ (not shown here).

### B. STDP-Comaptiable Spike Generation

The shape of action potential function $V_{spk}$ strongly influences the resulting STDP-learning function. A biological-like STDP pulse with exponential rising edges is difficult for circuit implementation. However, a bio-inspired STDP learning function can be achieved with a simpler action potential shape by implementing narrow positive pulse of large amplitude and a longer relaxing negative tail, which still keeps a STDP learning function very similar to its biological counterpart [2].

As shown in Fig. 5, we used a voltage selector with a RC charging circuitry to generate positive and negative tails. An on-chip configurable voltage reference was built in to control spike amplitude $V_a^+$ and $V_a^-$. In addition, digitally configurable capacitor and resistor banks were implemented to offer spike pulse tunability to optimize their response to a range of resistive synapse characteristics (e.g., threshold voltage and the program/erase pulse shape required by the spike-based learning algorithms [1]). Thanks to the dual-mode operation, two connected neurons can drive a pair of these spikes (pre- and post-) into the synapse between them directly. With difference in arriving time ($\Delta T$), pre- and post-synaptic spikes create net potential, $V_{net} = V_{post} - V_{pre}$, across the resistive synapse and modifies the weight if $V_{net}$ over the threshold $V_p$ or $V_n$.

A phase control circuit was designed to generate two non-overlapping control signals, $\Phi_{int}$ and $\Phi_{fire}$, switching the IFN between the two operation modes. Together with another two non-overlapping phase signals, $\Phi_1$ for positive tail and $\Phi_2$ for negative tail, they define the timing of spike generation.

### III. SIMULATION RESULTS

We designed all circuits in Cadence Virtuoso analog design environment, and ran simulations in Cadence Spectre simulator. We used IBM 180nm standard CMOS process for circuits' realization. In integration mode, the opamp has DC gain of 39dB, 3V/μs slew rate and 5MHz unit gain frequency; while in firing mode, it has DC gain of 60dB, 15MHz unit gain frequency and 15V/μs slew rate when accommodating up to 10,000 memristors described in [32] each has 1MΩ resistance. The STDP generator circuit was designed to be configurable to allow a broad range of memristors. Such tunability is also useful in physical circuits' implementation to compensate memristor character variations. We used a published device model [28] that has been matched to multiple physical memristors [29]–[33] and resistive random access memory characterizations [34] for memristor simulation. The model was coded with Verilog-A and device parameters matched to [32] were applied with $V_p$ = 0.16V and $V_n$ = 0.15V.

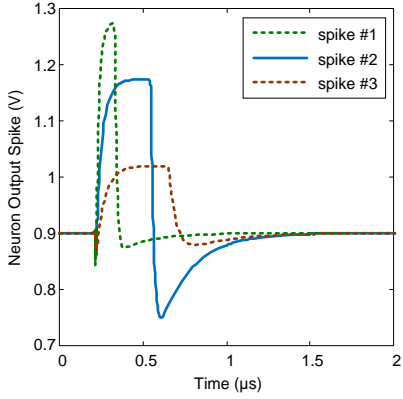

Fig. 6. Examples of neuron output spikes generated from the digitally configurable spike generator. The waveform shape of narrow and tall positive tail and wide slowly rising negative tail enables STDP.

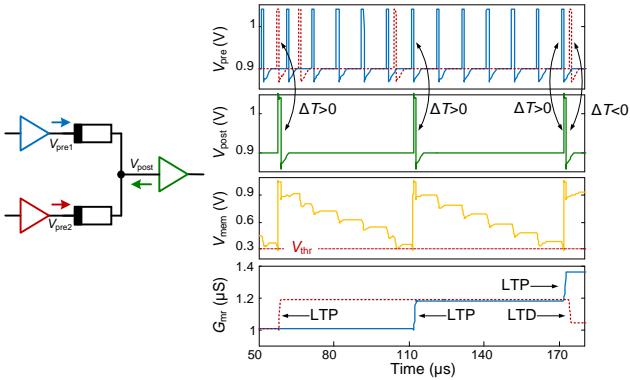

Fig. 7. The neurons' operation and STDP learning in a small system with two input neurons and one output neuron. Output neuron sums input current and yields $V_{mem}$. Post-synaptic spikes $V_{post}$ fired when $V_{mem}$ crosses $V_{th}$ caused synaptic potentiation or depression, which depends on the relative arriving time with respect to the pre-synaptic spikes $V_{pre}$.

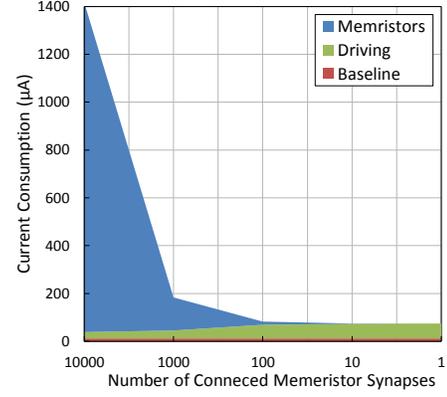

Fig. 8. Current consumption breakdown. Current proportional to synapse numbers were required to sustain spike voltage amplitudes for desired STDP learning in memristors, which causes large current being pulled when the number of synapses is large. Dynamic biasing based on dual-mode operation kept neuron very low power with only baseline current in integration mode, and extra current for output driving in firing mode.

Fig. 6 shows three examples of the output STDP spike generated from the configurable spike generator with positive/negative amplitudes and pulse widths were set to various values, while using 1.8V power supply and driving 1,000 memristor synapses with their resistance tightly distributed around 1MΩ. The shape of spike is adjustable to accommodate a broad range of memristor characteristics and the circuit behavior mandated by SNN learning algorithms.

STDP learning was tested in a small system with two memristor synapses were connected between two input neurons (pre-synaptic neurons) and one output neuron (post-synaptic neuron). As show in Fig. 7, one of the pre-synaptic neurons was forced to spike $V_{pre1}$ (solid line) regularly, while the other was spiking $V_{pre2}$ (dash line) randomly. The post-synaptic neuron summed currents converted from $V_{pre1}$ and $V_{pre2}$ by the two synapses, and yielded $V_{mem}$. Post-synaptic spikes $V_{post}$ were generated once $V_{mem}$ ran across a $V_{thr}$ = 0.3V. All spikes were set with the same parameters: $V_a^+$ = 140mV, $V_a^-$ = 30mV, $tail^+$ = 1μs and $tail^-$ = 3μs. The bottom panel of Fig.7 shows long-term potentiation (LTP) and long-term depression (LTD) of memristor synapses when post-synaptic spikes overlapped with the latest pre-synaptic spikes. Quantitatively, a post/pre-synaptic spike pair with 1μs arriving time difference $\Delta T$ resulted in a 0.2μS conductance increase or decrease depending on late or earlier arrival of $V_{post}$ relative to $V_{pre}$ respectively. It is worth to notice that the shape of the generated STDP spike was designed to be small enough to avoid perturbing memristor, at the same time, be large enough to be able create net potentials across memristor with potential above the programming thresholds of the memristors.

To evaluate the energy-efficiency, the neuron was designed to have a driving capability up to 10,000 memristor synapses each having 1MΩ resistance, which yields a 100Ω equivalent resistive load. Fig. 8 shows the neuron consumed 13μA baseline current in integration mode. When firing, the dynamically biased output stage consumed around 56μA current for driving, and passed the other current to memristor synapses: 1.4mA peak current for 10,000 memristor synapses to sustain the spike voltage amplitude of 140mV. The current sunk by the synapses simply follows Ohm's law due to the nature of memristor synapses as resistive-type load. Insufficient current supplied to memristors will cause lower spike voltage amplitude that may consequently lead to failure of STDP learning. Here, the widely used energy-efficiency merit for silicon neuron, *pJ/spike/synapse*, is not effective. Instead, the power efficiency $\eta$ during the maximum driving condition (at equivalent resistive load) should be used

$$\eta = \frac{I_{mr}}{I_{mr} + I_{IFN}}$$

where $I_{mr}$ is the current consumed by memristors and $I_{IFN}$ is the current consumed by silicon neuron. Our simulation demonstrated $\eta$ = 97% at 100Ω for the selected memristor, and the baseline power consumption of 22μW with a 1.8V power supply voltage, which means the CMOS neuron transferred most of the energy to memristors; while itself consumed only 3% of the energy efficiently to drive such a large number of synapses.

Finally, Table I shows the comparison results with the related works. It should be noted that most of previous silicon neuron designs don't accommodate two-terminal memristor, and therefore, it is inapplicable to compare the figures directly. While the best comparable works are the neurons reported in [2], [25]–[27], but unfortunately, they don't report the crucial power figures.

TABLE I. COMPARASION TABLE

|  | This Work | [25][27] | [2][26] | [10][21][22] |
|---|---|---|---|---|
| Memristor Compatible | × | × | × | No |
| Fixed $V_{refr}$ for Synapses | × | × | × | - |
| Current Summing Node | × | × | × | - |
| STDP-Compatible Pulse | × | × | × | - |
| Dynamic Powering | × | No | No | - |
| Baseline Power | 22μW | N/A[1] | N/A[1] | Vary[2] |
| Large Driving Current | × | No | No | - |
| Large Driving Efficiency | 97% | N/A[1] | N/A[1] | - |

1. The figure is not reported.
2. Inapplicable to compare.

## IV. DISCUSSION

The described CMOS spiking neuron architecture is generalized for memristor synapses. By selecting appropriate CMOS technology, online STDP learning can be achieved with memristors reported in [29], [30], [32]–[34]. However, the memristor in [31], with its $V_p$ = 1.5V and $V_n$ = 0.5V, has difficulties to fit into this architecture because the STDP pulse can produce both LTP and LTD while not disturbing memristor otherwise, doesn't exist. In other words, for generalized STDP learning, assuming pre- and post-synaptic spike are symmetric, needs a memristor synapse that has $|V_p - V_n| < \min(V_p, V_n)$.

In terms of energy-efficiency, an optimized design is the one with driving capability tailored according to desired application. For instance, widely used MNIST pattern recognition with single-layer perceptron needs 784 synaptic connections to each decision neuron, thus the average resistive loading of these 784 synapses should be evaluated in both training and testing scenarios. Then the neuron driving capability is selected to sustain the least spike voltage amplitudes on the lowest equivalent resistive load while achieving the highest power efficiency. In another case, e.g. 480×640 imaging patterns, a neuron with huge driving capability for 30,720 synapses may be required or alternative learning solution to cut the synaptic connections to a neuron is needed.

## V. CONCLUSION

This paper described a concise and yet elegant novel CMOS spiking integrate-and-fire neuron circuit for brain-like neuromorphic computing systems. The main strengths lie in its capability of driving a large number of memristor synapses, enabling online STDP learning and optimized energy-efficiency. Simulation results verified its functionality, shown up to 97% power efficiency when driving STDP learning in 10,000 memristor synapses with a nominal 1MΩ memristance, and the worst baseline power consumptions of 22μW for integration and 112μW for firing.

## VI. ACKNOWLEDGMENT

This work was supported in part by the National Science Foundation under the grant CCF-1320987.